\documentclass{article}
\usepackage{spconf,amsmath,graphicx}

\usepackage{booktabs}
\usepackage{multirow} 
\usepackage{subcaption}
\usepackage{float}

\title{Density-Guided Dense Pseudo Label Selection For Semi-supervised Oriented Object Detection}
\name{Tong Zhao \quad Qiang Fang{$^{\dag}$} \quad Shuohao Shi \quad Xin Xu \thanks{{$^{\dag}$}Corresponding Author, E-mail: qiangfang@nudt.edu.cn}}
  
\address{College of Intelligence Science and Technology, National University of Defense Technology, China }
\begin{document}
%
\maketitle
\begin{abstract}
Recently, dense pseudo-label, which directly selects pseudo labels from the original output of the teacher model without any complicated post-processing steps, has received considerable attention in semi-supervised object detection (SSOD). However, for the multi-oriented and dense objects that are common in aerial scenes, existing dense pseudo-label selection methods are inefficient because they ignore the significant density difference. Therefore, we propose Density-Guided Dense Pseudo Label Selection (DDPLS) for semi-supervised oriented object detection. In DDPLS, we design a simple but effective adaptive mechanism to guide the selection of dense pseudo labels. Specifically, we propose the Pseudo Density Score (PDS) to estimate the density of potential objects and use this score to select reliable dense pseudo labels. On the DOTA-v1.5 benchmark, the proposed method outperforms previous methods especially when labeled data are scarce. For example, it achieves 49.78 mAP given only 5\% of annotated data, which surpasses previous state-of-the-art method given 10\% of annotated data by 1.15 mAP. Our codes is available at https://github.com/Haru-zt/DDPLS.
\end{abstract}
\section{Introduction}
\label{sec:intro}

Recently, object detection has achieved great success with the help of sufficient labeled data. However, annotating abundant fully labeled datasets is a costly and time-consuming process. Therefore, to effectively leverage abundant unlabeled data, semi-supervised object detection (SSOD) has received extensive attention. The existing works in SSOD \cite{soft, zhou2022dense, wang2023consistent, liu2023mixteacher} mainly focus on general object detection, in which the objects are annotated with horizontal boxes. However, in some scenes, such as aerial images, horizontal boxes have difficulty efficiently representing objects \cite{roitransformer, xia2018dota}. In contrast to those in general scenes, objects in aerial images are typically captured from the bird's-eye view (BEV) and consequently present additional challenges including arbitrary orientations, small scales and dense distribution \cite{roitransformer}. Therefore, for the processing of such images, semi-supervised oriented object detection should be given serious consideration. 

\begin{figure}
	\centering
	\includegraphics[width=1.0\linewidth]{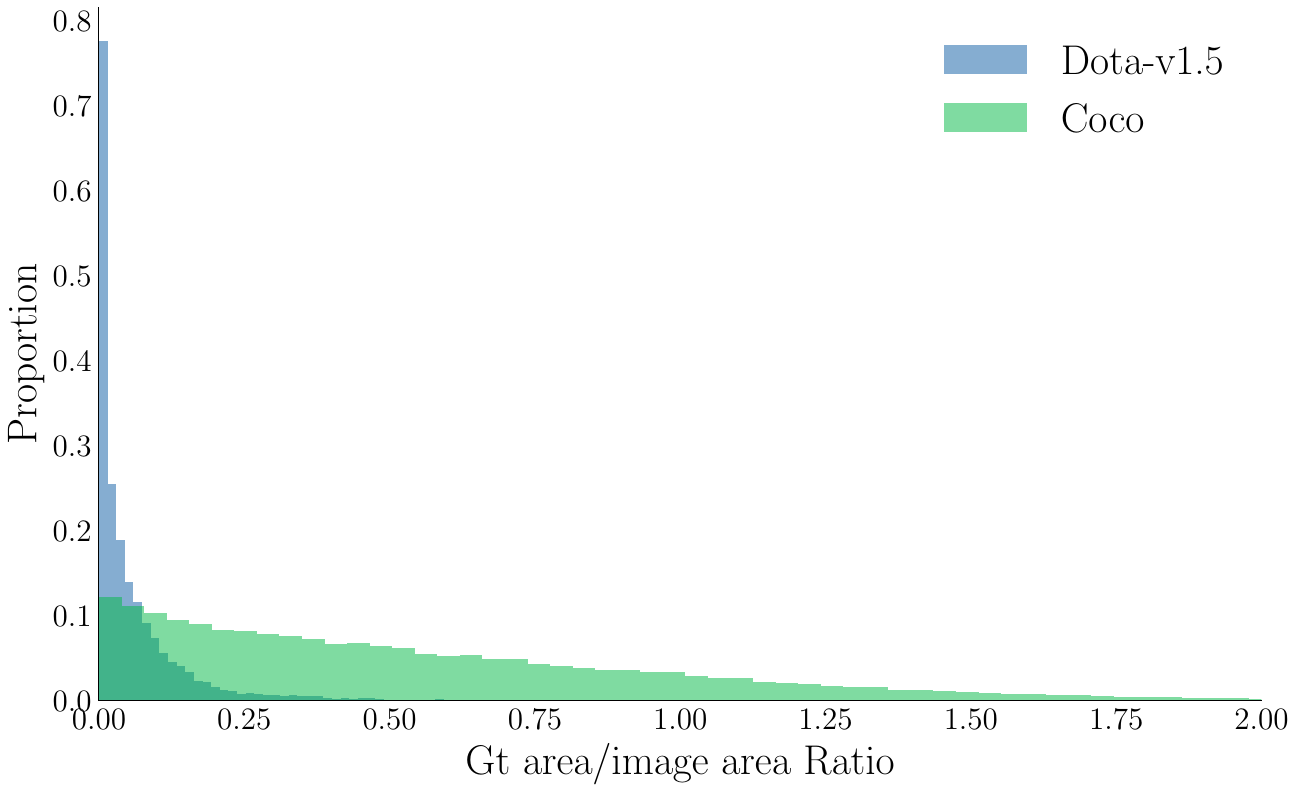}	\caption{Significant density difference in DOTA-v1.5 compared with COCO}
	\label{fig:ratio}
\end{figure}

Existing SSOD methods strongly rely on the precise pseudo labels, which can be divided into spare pseudo-label \cite{liu2021unbiased, soft, wang2023consistent} and dense pseudo-label \cite{zhou2022dense}, according to the sparsity of pseudo-label. In the spare pseudo-label, bounding boxes and their labels are provided as the supervision information, similar to the ground truth. And some strict conditions are applied to select reliable pseudo labels \cite{soft}. However, for the dense pseudo-label, pseudo labels are directly selected from the original output of the teacher model without any complicated post-processing steps. By removing post-processing steps, dense pseudo-label retains richer information \cite{zhou2022dense} and thus has received extensive attention. 

However, for aerial scenes, existing dense pseudo-label selection methods are inefficient. Dense Teacher \cite{zhou2022dense} proposes a region selection technique to highlight the key information and suppress noise but it requires a fixed selecting ratio to control the number of pseudo labels, which limits the ability to select sufficient pseudo labels in dense scenes, and may cause the selected pseudo labels to contain abundant noise in other scenes. SOOD \cite{hua2023sood} combines dense pseudo-label with spare pseudo-label to reduce noise. In SOOD \cite{hua2023sood}, dense pseudo labels are randomly sampled from the teacher's predictions but consequently involve a sequence of post-processing steps with fine-tuned hyper-parameters, which has been shown to be sensitive in dense scenes \cite{zhou2022dense}.  

In this study, we find that an important factor contributing to the above problems is that the density of potential objects is not taken into account in the existing dense pseudo-label selection methods. As shown in Fig \ref{fig:ratio}, in general scenes like COCO, objects tend to be evenly distributed and the importance of potential objects' density is ignored. However, in aerial scenes like DOTA-v1.5, objects tend to be densely distributed, which means that most of the objects are concentrated in a small area while the rest mainly consists of background. In this case, considering the density of potential objects during the selection of dense pseudo-label can greatly facilitate the selection process. 

Therefore, we propose Density-Guided Dense Pseudo Label Selection (DDPLS) for semi-supervised oriented object detection. The key component of DDPLS is an adaptive mechanism designed to estimate the density of potential objects in an image and use it to guide the selection of dense pseudo-label. Specifically, we consider that the post-sigmoid logit predictions of the teacher model can be act as indicators of where the features are rich \cite{zhixing2021FRS}. Thus we propose the Pseudo Density Score (PDS) to estimate the density of potential objects contained in an image. With the help of DDPLS, we formulate a direct way to integrate potential density information into the dense pseudo-label selection process.

The proposed DDPLS method greatly surpasses current semi-supervised oriented object detection method especially when labeled data are scarce. Specifically, our approach reaches 55.68 mAP with 10\% labeled data on the DOTA-v1.5 benchmark, surpassing the previous best method SOOD \cite{hua2023sood} by 7.05 points.
Concretely, this paper offers the following contributions to the field.

\begin{itemize}
\item We find that ignoring the potential objects' density in the existing dense pseudo-label selection methods impedes the performance of semi-supervised oriented object detection.
\item We propose a simple but effective method, called Density-Guided Dense Pseudo Label Selection \\(DDPLS), to formulate a direct way to integrate potential density information into the dense pseudo-label selection process and select suitable pseudo labels.
\item Our DDPLS method achieves state-of-the-art performance under various settings on the DOTA-v1.5 dataset.
\end{itemize}

\section{Related works}
\label{sec:rel}

\begin{figure*}
	\centering
	\includegraphics[width=1.0\linewidth]{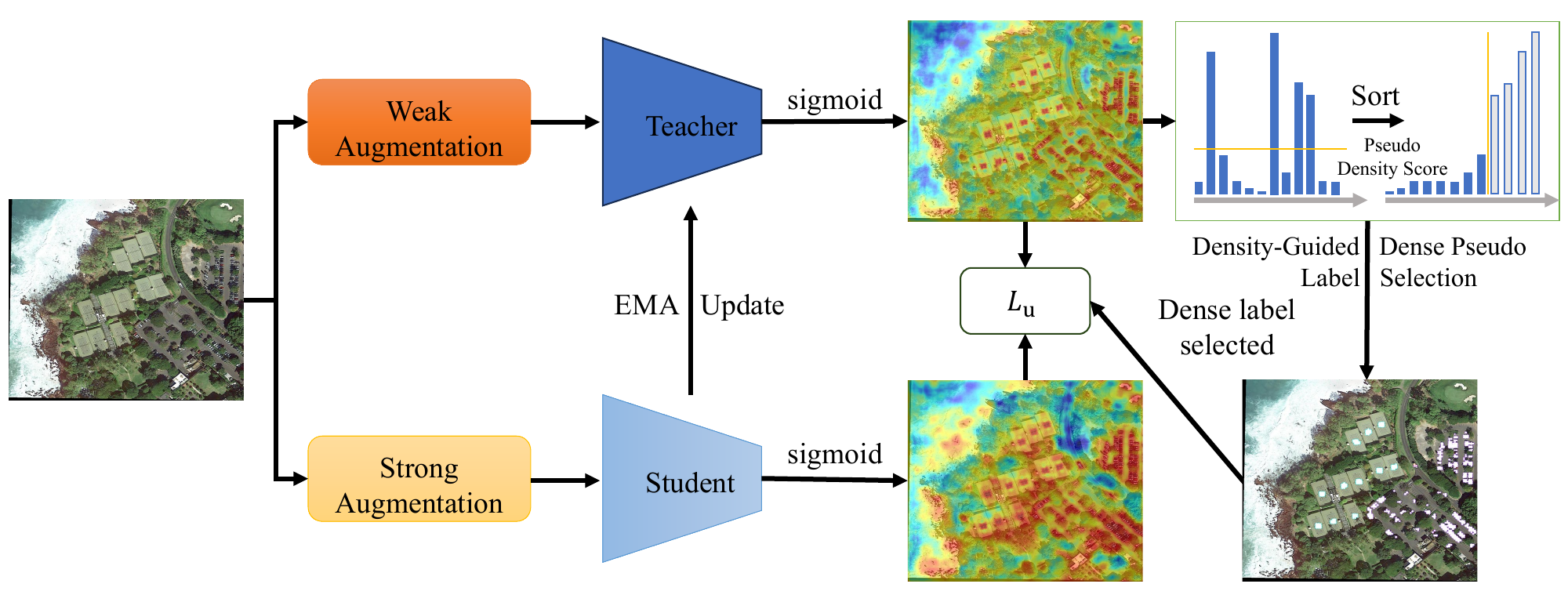}
	\caption{The overview of proposed DDPLS. Each training batch consists of both labeled data and unlabeled data. Note that we hide the supervised part for simplicity. For the unsupervised part, we sample dense pseudo labels according to the Density-Guided Dense Pseudo Label Selection.}
	\label{fig:overview}
\end{figure*}

\subsection{Semi-Supervised Object Detection}  
In the past few years, semi-supervised learning (SSL) \cite {mean,berthelot2019mixmatch, sohn2020fixmatch} has achieved significant improvements in image classification. Related methods can be roughly divided into consistency regularization \cite{mean, berthelot2019mixmatch} and pseudo-labeling \cite{lee2013pseudo}. In contrast to SSL, SSOD is more challenging. It involves instance-level predictions and bounding box regression. As an early attempt, CSD \cite{jeong2019CSD} uses horizontal flipping augmentation and relies on a consistency loss to constrain the model. STAC \cite{sohn2020simple} utilizes weak and strong augmentations inspired by FixMatch \cite{sohn2020fixmatch}. Later, researchers \cite{liu2021unbiased, soft} simplify the training procedure by adopting EMA from Mean Teacher \cite{mean} to realize end-to-end training. Soft Teacher \cite{soft} adopts an adaptive weight for each pseudo-box according to the classification score and uses the box jittering method to select reliable pseudo-boxs for the regression branch. In Dense Teacher \cite{zhou2022dense}, sparse pseudo-boxes are replaced with dense predictions as a straightforward form of pseudo-label to eliminate the influence of post-processing and handcrafted thresholds. Consistent Teacher \cite{wang2023consistent} is proposed by combining adaptive anchor assignment, 3D feature alignment module and Gaussian Mixture Model to reduce inconsistency during training. However, the above works all focus on general object detection. This paper aims to improve the performance of semi-supervised oriented object detection.

\subsection{Oriented Object Detection}
Different from general object detection, oriented object detection locates objects with Oriented Bounding Boxes (OBBs). Typically, oriented objects are captured from BEV perspective, as in the images acquired by satellites and unmanned aerial vehicles (UAVs). In recent years, many methods have been proposed to improve the performance in this area. RoI Transformer \cite{roitransformer} is proposed by combining an RRoI learner to convert horizontal regions of interest (HRoIs) into rotated regions of interest (RRoIs) and an RPS RoI Align module to extract spatially rotation-invariant feature maps. In ReDet \cite{han2021redet}, rotation equivariant networks and RiRoI Align are adopted to extract rotation-invariant features in both spatial and orientation dimensions. Oriented RCNN \cite{xie2021oriented} designs a lightweight module to generate oriented proposals and proposes a midpoint offset to represent objects. LSKNet \cite{LSKNet} explores large and selective kernel mechanisms in orient object detection. However, these works are all implemented in a supervised setting, whereas the present paper focuses on semi-supervised oriented object detection.


\subsection{Semi-supervised Oriented Object Detection}
Recently, SOOD \cite{hua2023sood} first explores semi-supervised oriented object detection by introducing global consistency and adaptive weights based on the orientation gap between the teacher and the student models, achieving excellent performance. SOOD \cite{hua2023sood} selects reliable dense pseudo labels by randomly sampling from the teacher's predictions, but thus it requires a sequence of post-processing with fine-tuned hyper-parameters. Compared with SOOD \cite{hua2023sood}, our method designs an adaptive mechanism to select appropriate dense pseudo labels based on the density of potential objects contained in an image, thus removing above problems.  

\section{Method}
\label{sec:method}

In this section, we introduce our method in detail. In \ref{3.1}, we first introduce the pseudo-labeling framework. In \ref{3.2}, we further describe the proposed Density-Guided Dense Pseudo Label Selection (DDPLS) method. Fig \ref{fig:overview} illustrates an overview of our proposed method.

\subsection{Pseudo-labeling Framework}
\label{3.1}

We first introduce the pseudo-labeling framework \cite{soft, zhou2022dense} that is widely used in SSOD. Our method follows the teacher-student paradigm \cite{mean}. In each training iteration, a batch composed of both labeled images and unlabeled images is sampled, where the sampling is controlled by sample ratio $s_{r}$. Then the teacher model generates pseudo labels on the unlabeled images and the student model takes both the pseudo labels in unlabeled images and the ground-truth in labeled images as its training targets. The training objective consists of a supervised loss and an unsupervised loss:
\begin{equation}
    \mathcal{L}_{s} = \frac{1}{N_{l}}\sum_{i}^{N_{l}}[\mathcal{L}_{cls}(T(I_{l}^{i}))+\mathcal{L}_{reg}(T(I_{l}^{i})]
\end{equation}
\begin{equation}
    \mathcal{L}_{u} = \frac{1}{N_{u}}\sum_{i}^{N_{u}}[\mathcal{L}_{cls}(T^{\prime}(I_{u}^{i}))+\mathcal{L}_{reg}(T^{\prime}(I_{u}^{i})]
\end{equation}
where $\mathcal{L}_{cls}$ is the classification loss, $\mathcal{L}_{reg}$ is the regression loss, and $N_{l}$ and $N_{u}$ denote the numbers of labeled and unlabeled images, respectively. $I_{l}^{i}$ and $I_{u}^{i}$ represent the $i$-th labeled image and the $i$-th unlabeled image, respectively. $T$ and $T^{\prime}$ indicate weak and strong augmentation, respectively.
The overall loss is defined as a weighted sum of the supervised and unsupervised losses:
\begin{equation}
\mathcal{L} = \mathcal{L}_{s} + \alpha \mathcal{L}_{u}
\end{equation}

As training progresses, the parameters of the teacher model are updated in an EMA manner:

\begin{equation}
\theta_{t} = (1-\lambda)\theta_{t} + \lambda\theta_{s}
\end{equation}
where $\lambda$ is momentum.

\subsection{Density-Guided Dense Pseudo Label Selection}
\label{3.2}

The performance of the detector depends on the quality of the pseudo labels. To effectively leverage potential objects in the unlabeled data with significant density difference, we propose Density-Guided Dense Pseudo Label Selection (DDPLS) to integrate potential density information into the dense pseudo-label selection process.

Ideally, we want to select dense pseudo labels according to the really density of potential objects contained in an image. However, in practice, an accurate estimate of the potential object is impossible. Therefore, we seek an approximate method to estimate it. We consider that the post-sigmoid logit predictions can well serve as indicators of where the features are rich \cite{zhixing2021FRS} and are easily obtained in the pseudo-labeling framework. We empirically find that although not locally accurate, such feature density can be used as a global proxy to estimate the density of potential objects.

Specifically, we propose the Pseudo Density Score to estimate the density of potential objects. We define the Pseudo Density Score as:
\begin{equation}
    S_{PDS} = \frac{1}{N} \sum_{l=1}^{M} \sum_{i=1}^{W_l} \sum_{j=1}^{H_l} S_{lij}
\end{equation}
\begin{equation}
    S_{lij} = \mathop{\max}_{c} y_{lij,c}
\end{equation}
where $y_{lij,c}$ is the probability of category $c$ in the $l$-th feature pyramid network (FPN) layer at location ($i,j$) in the corresponding feature map. $M$ is the number of FPN layers, and $N=\sum_{l=1}^{M}\sum_{i=1}^{W_l}\sum_{j=1}^{H_l}1_{lij}$. 

With the help of the Pseudo Density Score, we can dynamically select the pixels in the top $\beta S_{PDS}\%$ as reliable dense pseudo label and the rest will be suppressed to 0 following the approach used in  Dense Teacher \cite{zhou2022dense}. As a result, the dense pseudo labels are selected according to:

\begin{equation}
    \Vec{y_{lij}} = \begin{cases}
        1, {\rm if} \quad S_{lij} \quad {\rm in} \quad {\rm top} \quad \beta S_{PDS}\%,\\
        0, {\rm otherwise}\\
    \end{cases}
\end{equation}
where $\Vec{y_{lij}}$ is the symbol deciding the selection of a dense pseudo label on the $l$-th FPN layer at location ($i,j$) in the corresponding feature map, and $\beta$ is a hyper-parameter for adjusting the intensity of selection, which we empirically set it to 100.   

\begin{figure}
	\centering
        \includegraphics[width=1.0\linewidth]{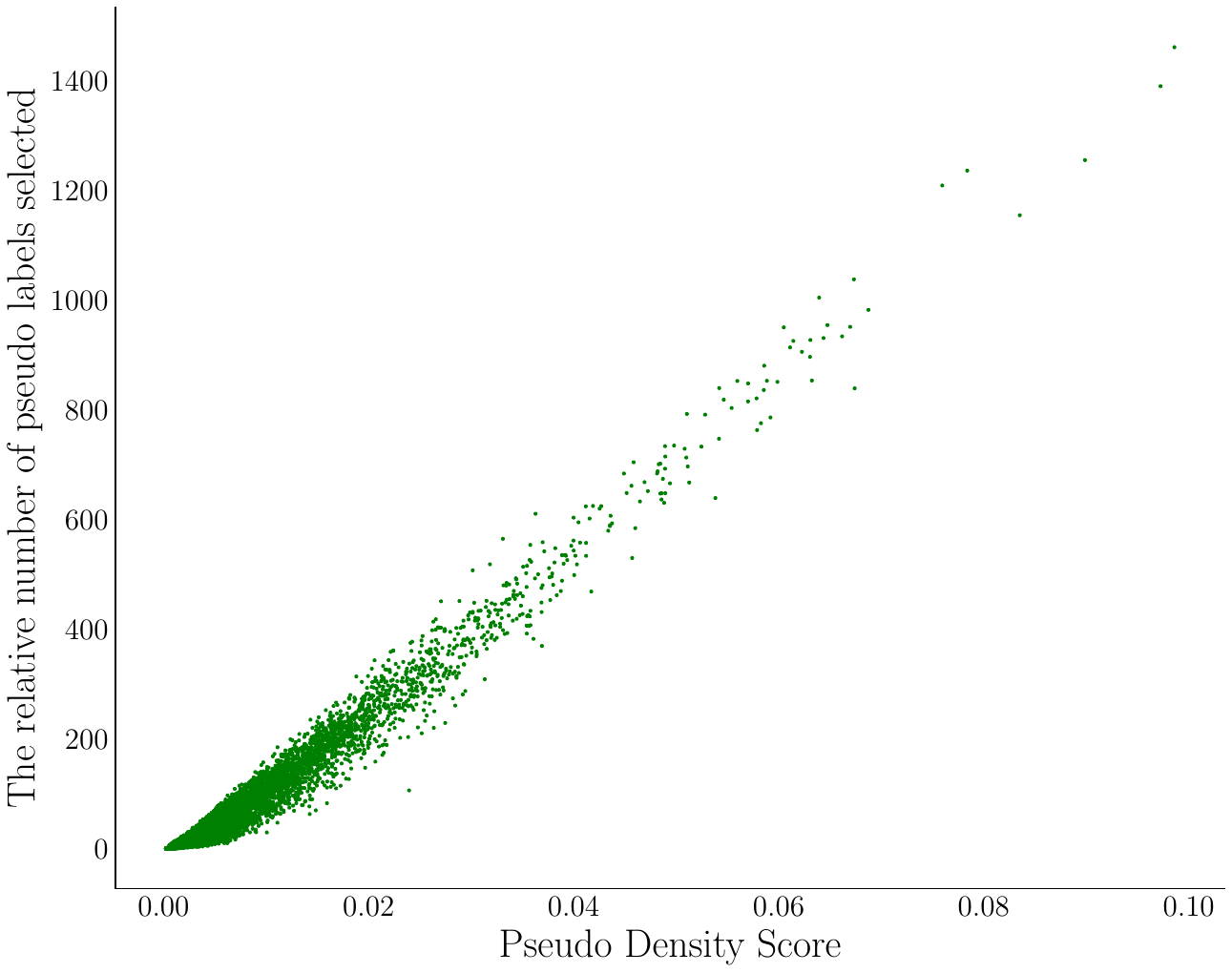}
        \caption{The correlation between the Pseudo Density Score and the relative number of pseudo labels selected under the 10\% setting. Relative number indicates the sum of confidence of pseudo labels selected.}
	\label{fig: fsr_num}
\end{figure}

\begin{figure}
	\centering
	\includegraphics[width=1.0\linewidth]{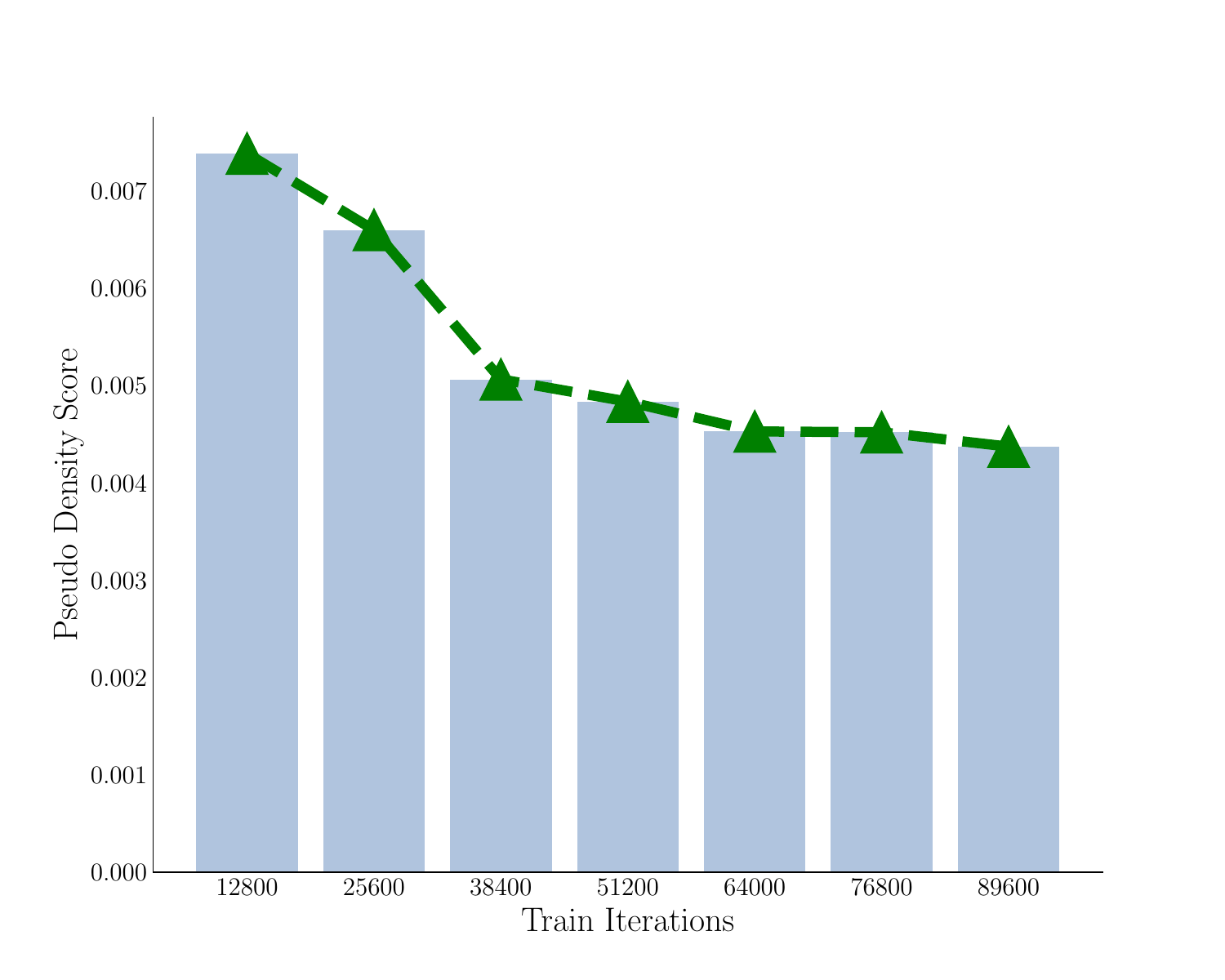}
	\caption{We randomly sampled 1k unlabeled training images under the 10\% setting and calculated the average Pseudo Density Score under different train iterations.}
	\label{fig:mean fsr}
\end{figure}\textbf{}
The empirical study in Fig \ref{fig: fsr_num} provides a demonstration of our hypothesis. The Pseudo Density Score has a positive correlation with the number of pseudo labels selected. Meanwhile, as shown in Fig \ref{fig:mean fsr}, as training progresses, the Pseudo Density Score gradually drops. We regard this phenomenon as indicative of a good dynamic semi-supervised train scheme because it is usually desirable to mine more potential information at the beginning of training to speed up convergence and then select increasingly accurate pseudo labels as training progresses to alleviate confirmation bias \cite{wang2023freematch}.

\section{Experiments}

\begin{table*}
  \centering
    \caption{Experimental results on DOTA-v1.5 under Parially Labeled Data setting. Experiments are conducted under 10\%, 20\% and 30\% labeled data settings. ${\ast}$ indicates reported performance in SOOD. Note that all methods are implemented with rotated-FCOS.}
  \begin{tabular}{@{}cccccccc@{}}
    \toprule
    Setting & Method& 1\% &5\% & 10\% & 20\% & 30\% \\
    \midrule
    \multirow{1}{*}{Supervised}
     & FCOS \cite{tian2019fcos}  & 15.67&36.18& 42.78 & 50.11 & 54.79\\ 
    \hline
     \multirow{3}{*}{Semi-supervised}
     &Dense Teacher \cite{zhou2022dense} $^{\ast}$&- &- & 46.90&53.93&57.86&\\
     &SOOD \cite{hua2023sood} &- & - &48.63&55.58&59.23\\
     &Ours &\textbf{29.95} &\textbf{49.78} &\textbf{55.68} &\textbf{60.98} &\textbf{62.59}\\
    \bottomrule
  \end{tabular}

  \label{tab:main-protortions}
\end{table*}

\begin{table}
  \centering
  \caption{Experimental results on DOTA-v1.5 under the Fully Labeled Data setting. Numbers in font of the arrow indicate the supervised baseline. ${\ast}$ indicates reported performance in SOOD. Note that all methods are implemented with rotated-FCOS.}
  \begin{tabular}{@{}ccc@{}}
    \toprule
     Method  & mAP \\
    \midrule
    Dense Teacher \cite{zhou2022dense} $^{\ast}$& $65.46\stackrel{+0.92}{\longrightarrow}66.38$ \\
    SOOD \cite{hua2023sood} & $65.46\stackrel{+2.24}{\longrightarrow}67.70$ \\
    Ours  & $65.44\stackrel{+2.79}{\longrightarrow}\textbf{68.23}$ \\
    \bottomrule
  \end{tabular}
  \label{tab:main-full}
\end{table}

\subsection{Dataset and Evaluation Protocol}

For a fair comparison, we validate our method on the DOTA-v1.5 benchmark following the previous work \cite{hua2023sood}. In DOTA-v1.5, two datasets are provided for model training: the DOTA-v1.5-train set contains 1411 labeled images, and the DOTA-v1.5-test set contains 937 unlabeled images. In addition, the DOTA-v1.5-val set with 458 labeled images is also provided for validation. In the previous study \cite{hua2023sood}, two settings were used for performance validation:

\noindent{\bf Partially Labeled Data.} SOOD \cite{hua2023sood} first introduced this setting. In this setting, 10\%, 20\% and 30\% of the images in the DOTA-v1.5-train set are sampled as the labeled training data, and the remaining unsampled images are treated as unlabeled data. For each protocol, one fold following the data distribution of the DOTA-v1.5-train is provided. To evaluate our method in more severe situations, we extend this setting to 1\% and 5\%. Note that in 1\% setting, only 14 images are provided as labeled data. 

\noindent{\bf Fully Labeled Data.} In this setting, the entire DOTA-v1.5-train set is used as the labeled data and the DOTA-v1.5-test set is used as the additional unlabeled data.

We evaluate our method under both settings and report the performance on DOTA-v1.5-val with the stand mean average precision (mAP) as the evaluation metric.

\subsection{Implementation Details}

We use FCOS \cite{tian2019fcos} equipped with a FPN as our default detection framework to evaluate the effectiveness of our method, and ResNet-50 pre-trained on ImageNet is adopted as the backbone. Our implementation and hyper-parameters are based on MMRotate \cite{zhou2022mmrotate}. Following the previous work \cite{hua2023sood}, we crop the original images into 1024 $\times$ 1024 patches with an overlap of 200.

\noindent{\bf Partially Labeled Data.} The model is trained for 120k iterations on 2 GPUs with 3 images per GPU. SGD is used, with the learning rate initialized to 0.0025. The weight decay and momentum are set to 0.0001 and 0.9, respectively. For fair comparison, we set the data sample ratio between the labeled and unlabeled data to 2:1 following the setting in SOOD \cite{hua2023sood}.

\noindent{\bf Fully Labeled Data.} The model is trained for 180k iterations on 2 GPUs with 3 images per GPU. SGD is used, with the learning rate initialized to 0.0025 and divided by 10 after 120k and 160k iterations. The other settings is same with Partially Labeled Data.

To maintain hard negative samples \cite{zhou2022dense}, we set the value of $\beta$ in DDPLS to 100 for both partially labeled data and fully labeled data. We adopt the data augmentation used in \cite{zhou2022dense}. Specifically, we use strong augmentation for the student model and weak augmentation for both the teacher model and supervised pipeline. Scale jittering and random flipping are adopted as weak augmentation, while the strong augmentation includes scale jittering, random flipping, color jittering, random grayscale, and random Gaussian blurring. Note that we do not apply random erasing used to avoid injecting excessively strong noise during training. Following previous works in SSOD \cite{liu2021unbiased, hua2023sood, wang2023consistent}, we adopt a "burn-in" strategy to initialize the teacher model. 

\subsection{Main Results}

In this section, we compare our method with previous state-of-the-art methods on DOTA-v1.5. 

\noindent{\bf Partially Labeled Data.} We evaluate our method under different proportions of labeled data , and the results are shown in Tab \ref{tab:main-protortions}. Our DDPLS method achieves state-of-the-art performance in all cases. Specifically, it obtains 55.68, 60.98, 62.59 mAP with 10\%, 20\%, and 30\% labeled data, surpassing the previous state-of-the-art method by +7.05, +5.40, and +3.36 points, respectively. Additionally, when we further evaluate our method with even fewer labeled data, it obtains 29.95 and 49.78 mAP on 1\% and 5\%, surpassing the supervised baseline by +14.28 and +13.60 points. Note that in the 5\% setting, our method achieves better performance (with only half of the labeled images) than SOOD \cite{hua2023sood} in 10\% setting.

The qualitative results of our method compared with supervised baseline and SOOD \cite{hua2023sood} are shown in Fig 5. With the help of DDPLS, the density of potential object can be exploited to guide the selection of dense pseudo label, helping obtain more information and improve the detection quality.

\noindent{\bf Fully Labeled Data.} We also compare our method with the previous state-of-the-art methods in the fully labeled data setting and the results are shown in Tab \ref{tab:main-full}. Since the reported performance of supervised baseline is different, we also report the performance of the supervised baseline. Our DDPLS method surpasses the previous state-of-the-art method by 0.53 points. Compared with the baseline, we achieve a +2.79 mAP improvement. 

\begin{table*}
  \centering
  \caption{Ablation study on different augmentations. Experiments are conducted at 10\%, 20\% and 30\% settings. ${\ast}$ and ${\dag}$ indicate without and with scale jittering, a common augumentation used in SSOD.}
  \begin{tabular}{@{}cccccc@{}}
    \toprule
    Setting & Method  & 10\% & 20\% & 30\% \\
    \midrule
    \multirow{1}{*}{Supervised}
    &FCOS \cite{tian2019fcos} & 42.78 & 50.11 & 54.79\\ 
    \hline
    \multirow{3}{*}{Semi-supervised$^{\ast}$}
    &Dense Teacher \cite{zhou2022dense} &46.90&53.93&57.86&\\
    &SOOD \cite{hua2023sood} &48.63&55.58&\textbf{59.23}&\\
    &Ours &\textbf{51.00} &\textbf{57.12} &59.00&\\
    \hline
    \multirow{3}{*}{Semi-supervised$^{\dag}$}
    &Dense Teacher \cite{zhou2022dense} &52.25&58.22 & 60.19&\\
    &SOOD \cite{hua2023sood} &51.39&58.05 & 60.81&\\
    &Ours &\textbf{55.68} & \textbf{60.98}&\textbf{62.59}&\\
     
    \bottomrule
  \end{tabular}
  
  \label{tab:ablation-aug}
\end{table*}

\begin{figure*}
  \centering
  \begin{subfigure}{0.21\textwidth}
    \includegraphics[width=1.0\linewidth]{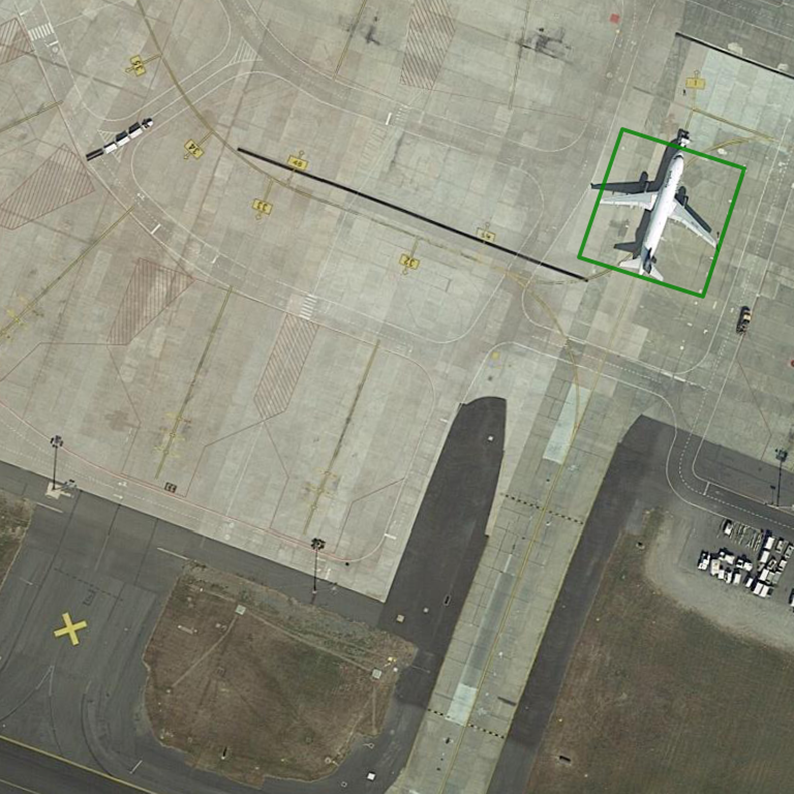}
  \end{subfigure}
  \begin{subfigure}{0.21\linewidth}
    \includegraphics[width=1.0\linewidth]{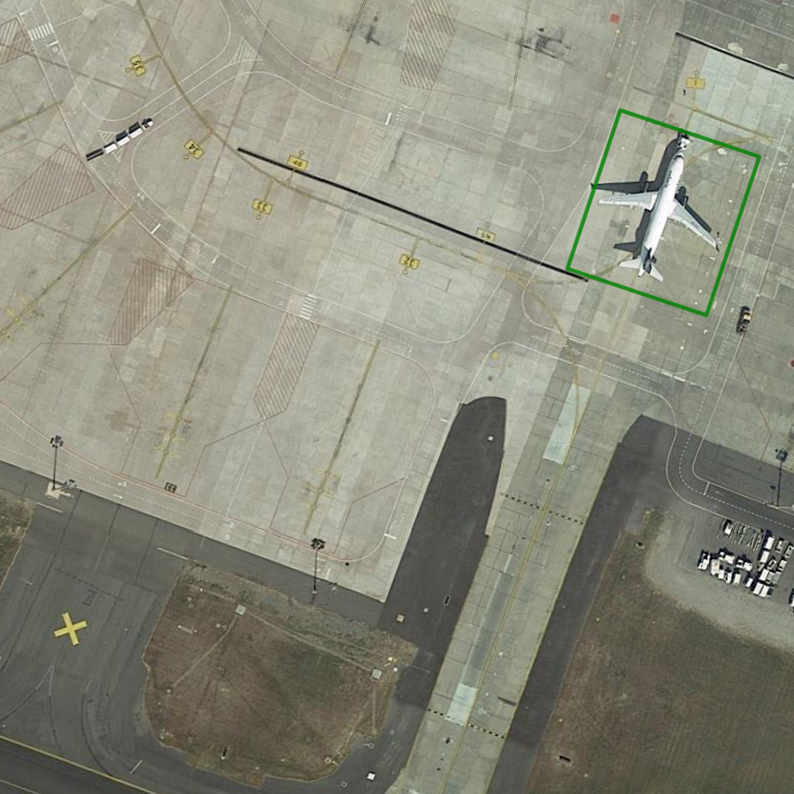}
  \end{subfigure}
  \begin{subfigure}{0.21\linewidth}
    \includegraphics[width=1.0\linewidth]{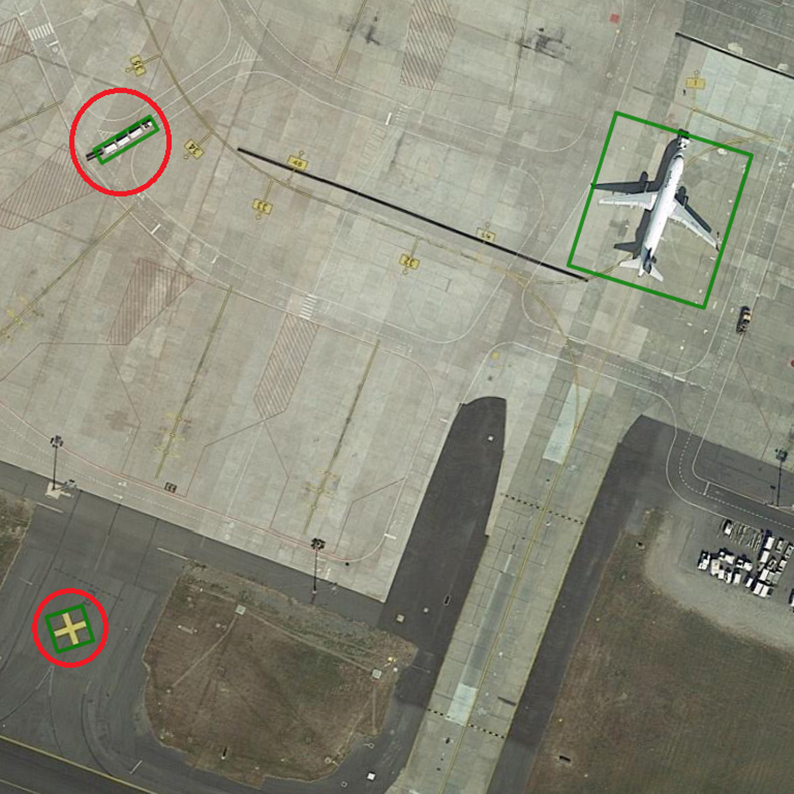}
  \end{subfigure}
  \begin{subfigure}{0.21\linewidth}
    \includegraphics[width=1.0\linewidth]{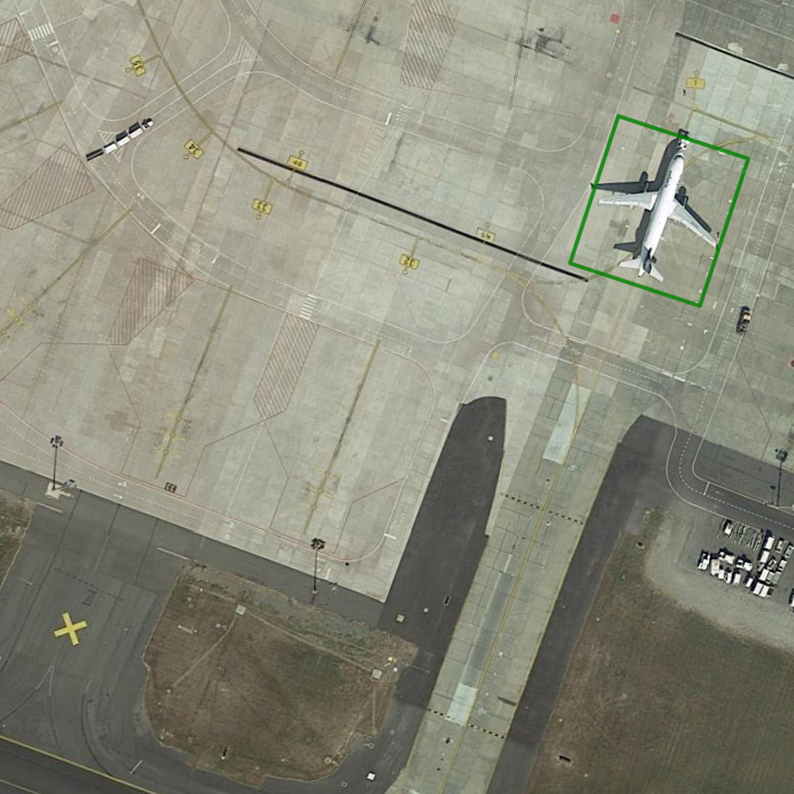}
  \end{subfigure}

  \begin{subfigure}{0.21\linewidth}
    \includegraphics[width=1.0\linewidth]{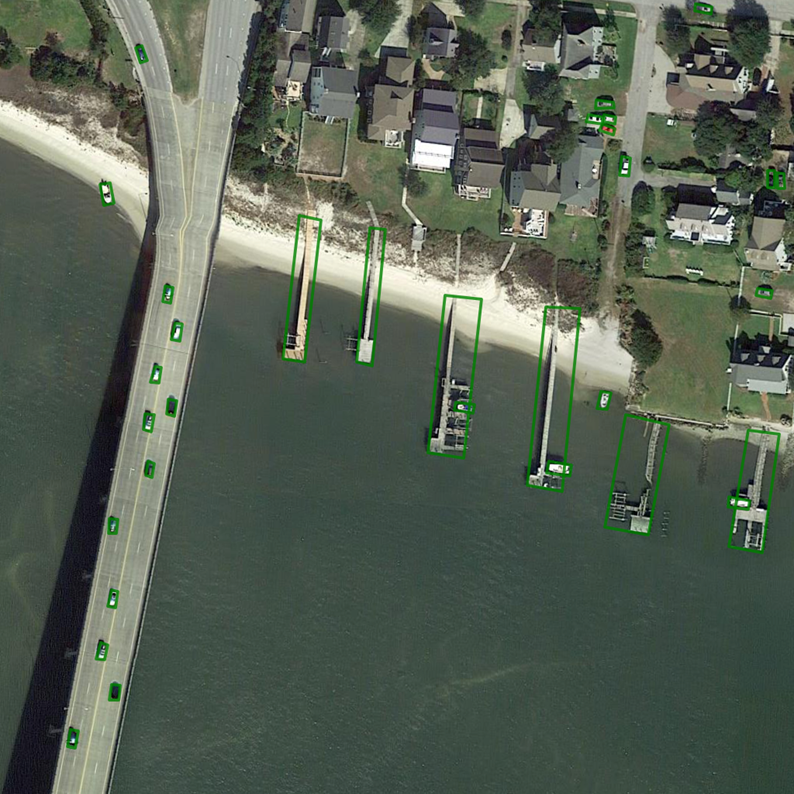}
  \end{subfigure}
  \begin{subfigure}{0.21\linewidth}
    \includegraphics[width=1.0\linewidth]{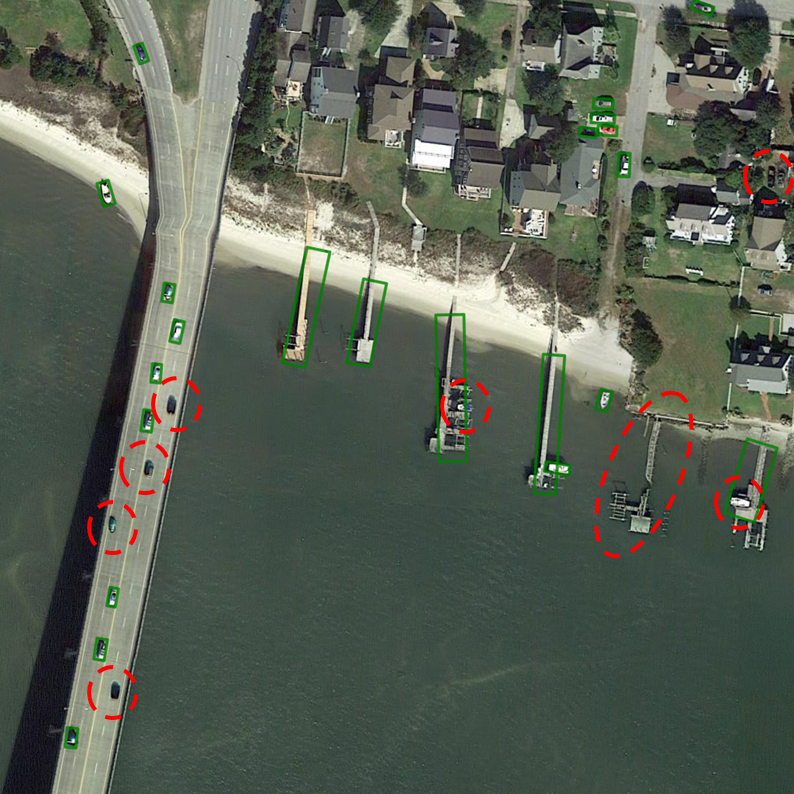}
  \end{subfigure}
  \begin{subfigure}{0.21\linewidth}
    \includegraphics[width=1.0\linewidth]{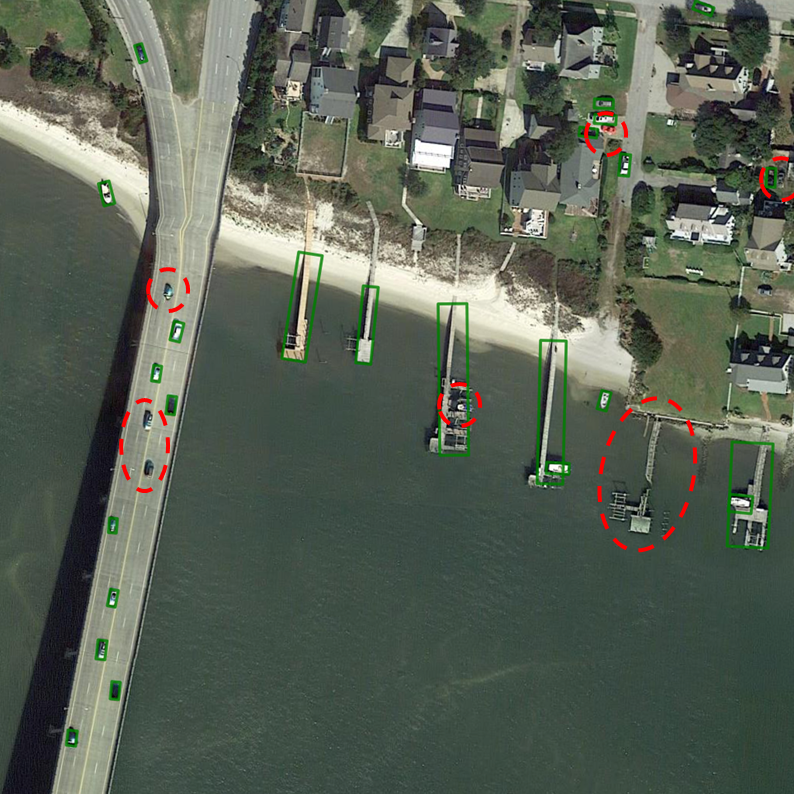}
  \end{subfigure}
  \begin{subfigure}{0.21\linewidth}
    \includegraphics[width=1.0\linewidth]{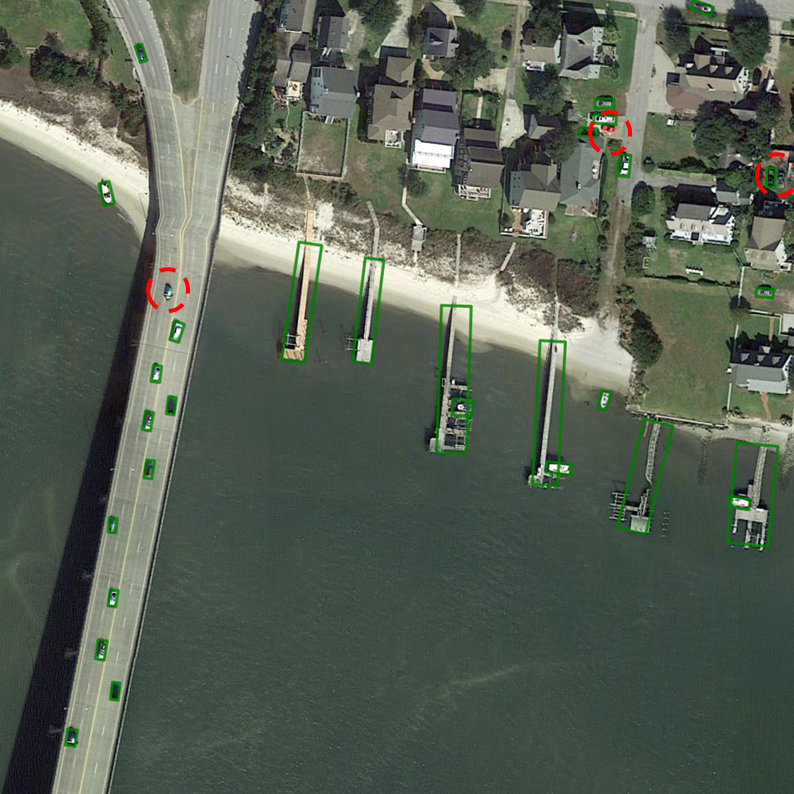}
  \end{subfigure}

  \begin{subfigure}{0.21\linewidth}
    \includegraphics[width=1.0\linewidth]{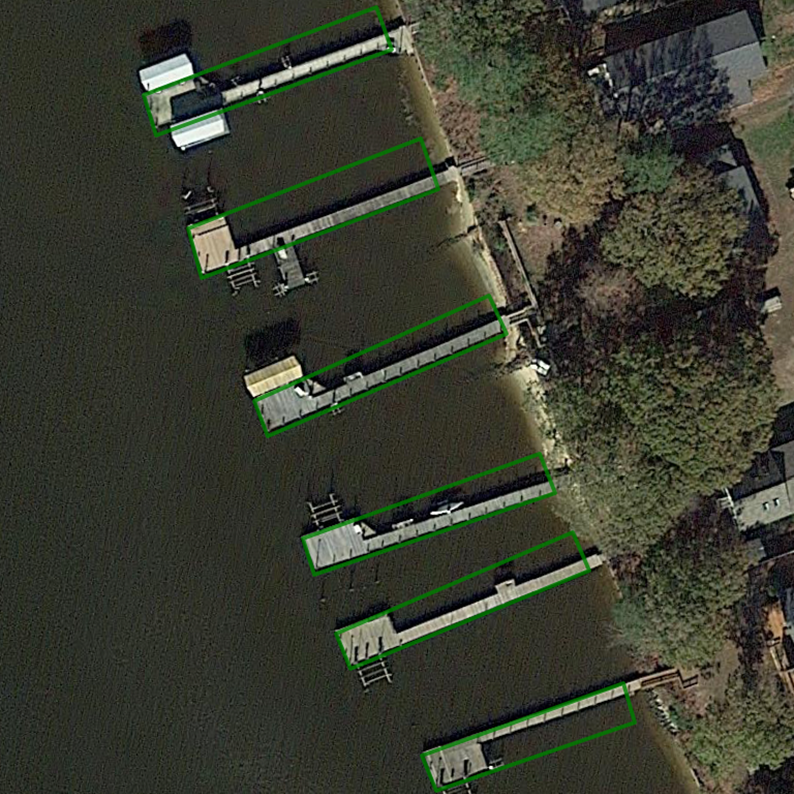}
  \end{subfigure}
  \begin{subfigure}{0.21\linewidth}
    \includegraphics[width=1.0\linewidth]{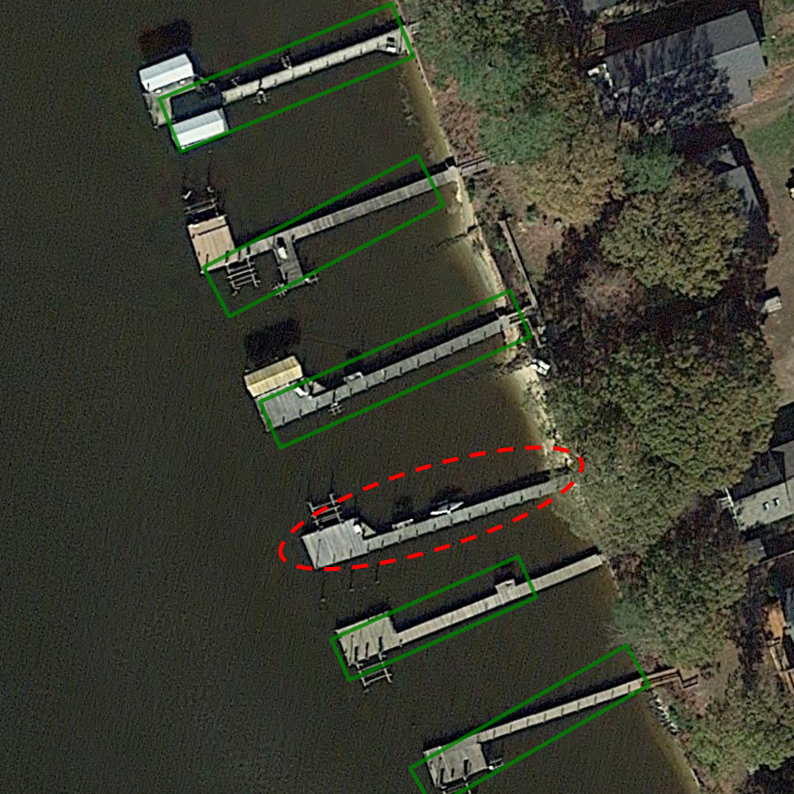}
  \end{subfigure}
  \begin{subfigure}{0.21\linewidth}
    \includegraphics[width=1.0\linewidth]{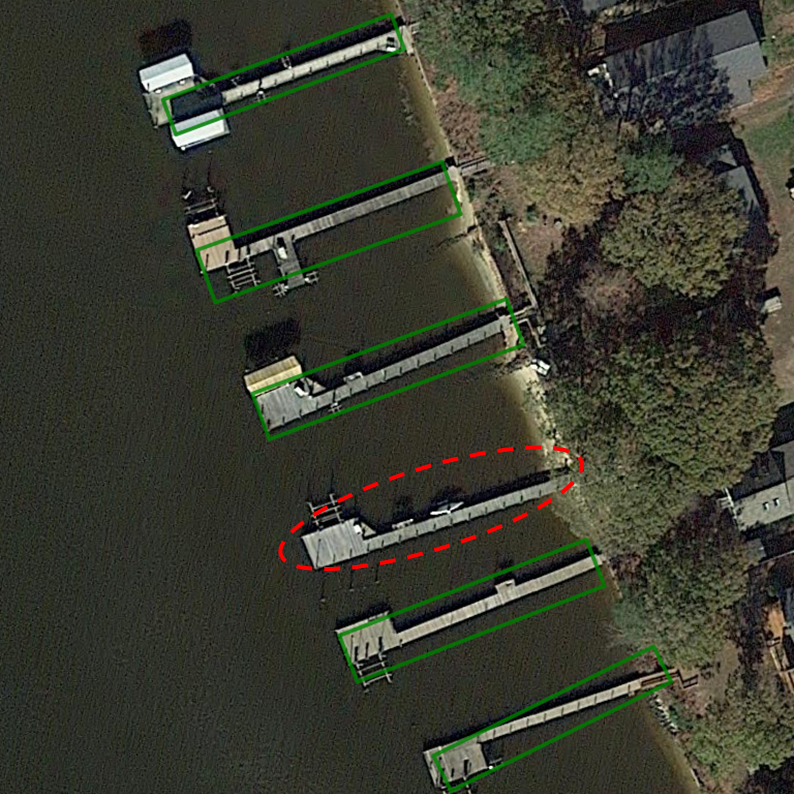}
  \end{subfigure}
  \begin{subfigure}{0.21\linewidth}
    \includegraphics[width=1.0\linewidth]{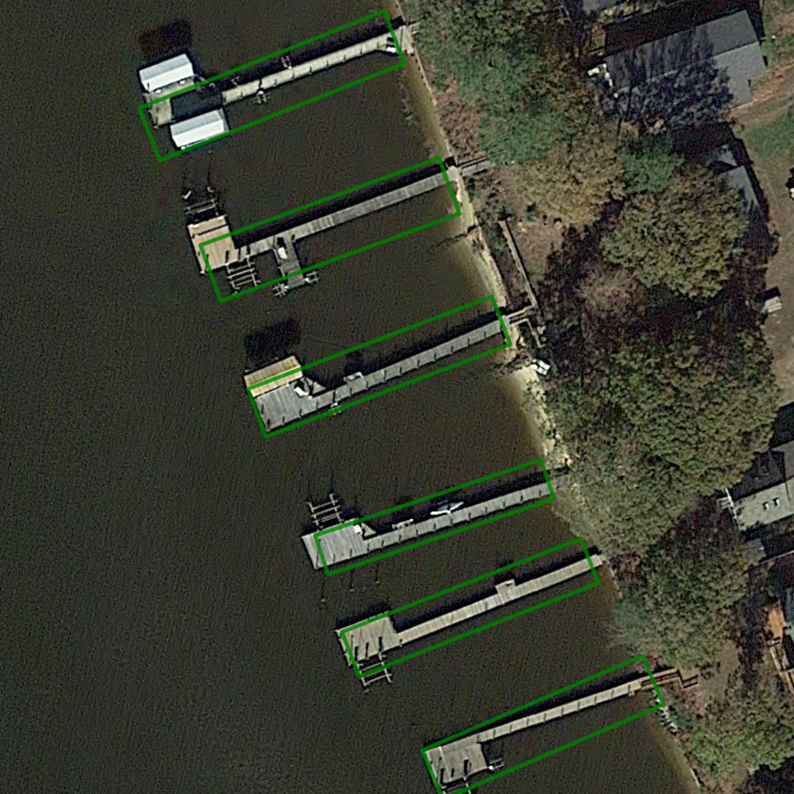}
  \end{subfigure}

  \begin{subfigure}{0.21\linewidth}
    \includegraphics[width=1.0\linewidth]{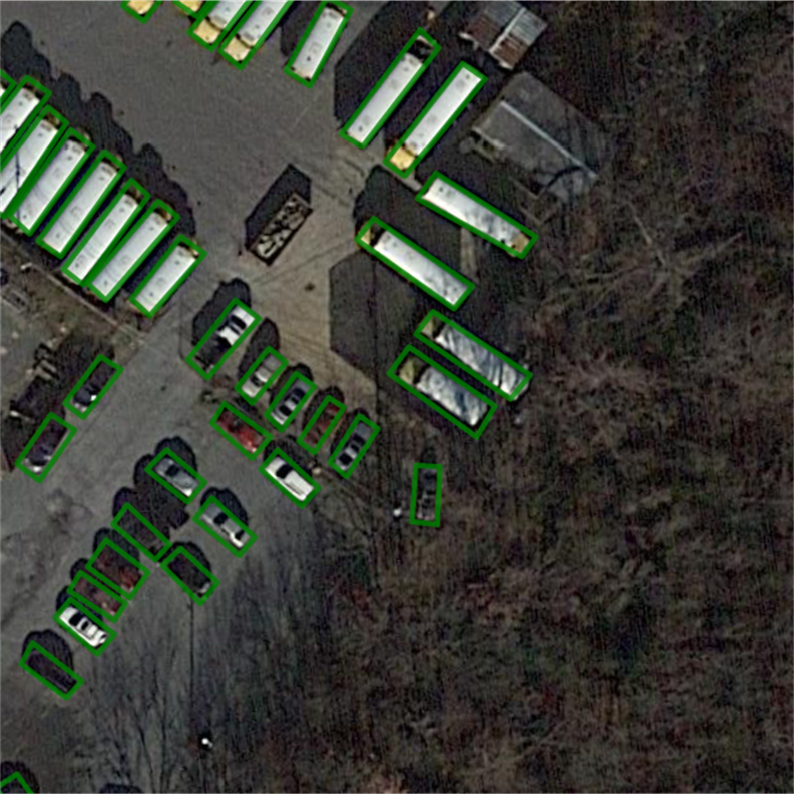}
    \caption{Ground Truth}
  \end{subfigure}
  \begin{subfigure}{0.21\linewidth}
    \includegraphics[width=1.0\linewidth]{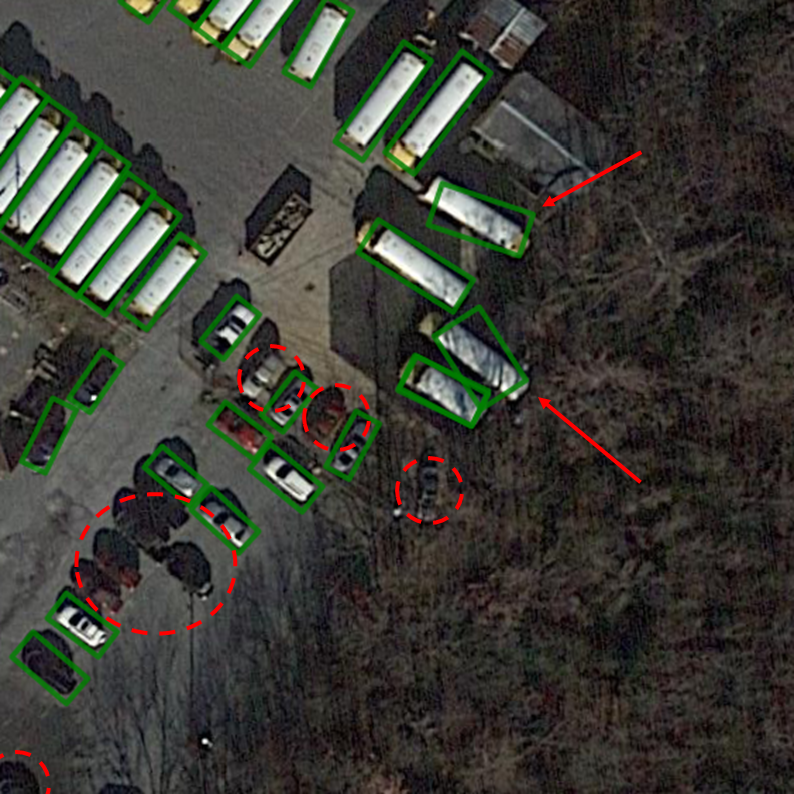}
    \caption{Supervised Baseline}
  \end{subfigure}
  \begin{subfigure}{0.21\linewidth}
    \includegraphics[width=1.0\linewidth]{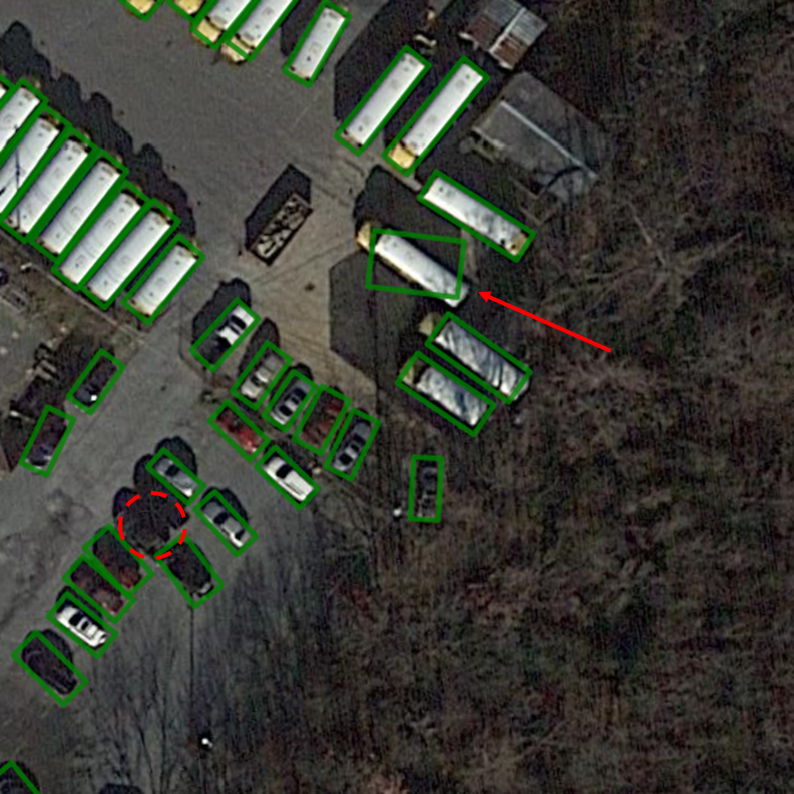}
    \caption{SOOD}
  \end{subfigure}
  \begin{subfigure}{0.21\linewidth}
    \includegraphics[width=1.0\linewidth]{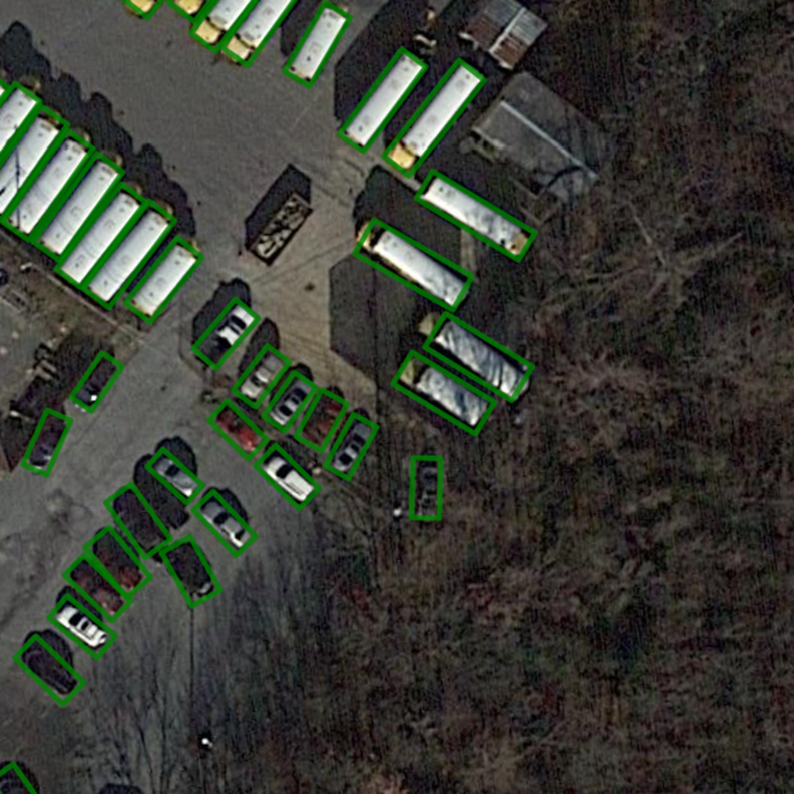}
    \caption{Ours}
  \end{subfigure}
  
  \caption{Some visualization examples from DOTA-v1.5 dataset. The green rectangles indicate predictions. The red dashed circle, solid red circle, and red arrow represent false negative, false positive, and inaccurate orientation prediction, respectively. }
  \label{fig:visualization examples}
\end{figure*}

\begin{table}
  \centering
  \caption{Ablation study on the effect of hyper-parameter $\beta$ in DDPLS. Experiments are conducted at 10\% setting.}
  \begin{tabular}{@{}ccc@{}}
    \toprule
    Setting  & $\beta$ & mAP \\
    \midrule
    \uppercase\expandafter{\romannumeral1} &10&47.09\\
    \uppercase\expandafter{\romannumeral2} &50&52.47\\
    \uppercase\expandafter{\romannumeral3} &100&\textbf{55.68} \\
    \uppercase\expandafter{\romannumeral4} &300&55.23\\
    \uppercase\expandafter{\romannumeral5} &500&55.54 \\
    \uppercase\expandafter{\romannumeral6} &700&55.04 \\
    \bottomrule
  \end{tabular}
  \label{tab:ablation-beta}
\end{table}

\begin{table}
  \centering
  \caption{Time cost analysis. Experiments are conducted at 10\% setting.}
  \begin{tabular}{@{}ccc@{}}
    \toprule
    Setting  & Method & Seconds \\
    \midrule
    \multirow{1}{*}{Supervised}
    &FCOS&0.20\\
    \hline
    \multirow{3}{*}{Semi-supervisd}
    &Dense Teacher&0.61\\
    &SOOD&0.59\\
    &Ours&0.64\\
    \bottomrule
  \end{tabular}
  \label{tab:ablation-time cost}
\end{table}
\subsection{Ablation Study}

\noindent{\bf The effect of augmentation.} In this part, we discuss the effect of different augmentations. We note that a weaker augmentation is adopted in SOOD. Specifically, we additionally use scale jittering to both labeled data and unlabeled data, as done in Soft Teacher \cite{soft} and Dense Teacher \cite{zhou2022dense}. Therefore, to determine the cause of the improved performance, we compare our method under the same augmentation in SOOD and re-implement SOOD in our augmentation. Moreover, Dense Teacher \cite{zhou2022dense}, which uses dense pseudo labels, is also evaluated for comparison. As shown in Tab \ref{tab:ablation-aug}, our method achieves 51.00 mAP and 57.12 mAP under the same augmentations of SOOD \cite{hua2023sood}, surpassing SOOD by +2.37 and +1.84 mAP. However, in the 30\% setting, our method achieves the performance of 59.00 mAP, which is 0.23 mAP behind the SOOD \cite{hua2023sood}. We conjecture that as the proportion of labeled data increases, the disadvantage of using sparsity pseudo labels to generate dense pseudo labels in SOOD is greatly compensated. 

When SOOD is re-implemented using the same augmentation used in DDPLS, we observe that our method outperforms SOOD in all settings, despite the fact that stronger data augmentation lead to performance gains in all methods. This confirms the validity of our approach. 

\noindent{\bf The effect of hyper-parameter $\beta$ in DDPLS.} Here, we study the influence of the hyper-parameter $\beta$ in DDPLS, which is the only hyper-parameter used. As shown in Tab \ref{tab:ablation-beta}, when we set $\beta$ to 10, we achieve the performance of 47.09 mAP. As $\beta$ increases from 10 to 100, the performance of our method improves. However, further increasing $\beta$ to 300 slightly hurts the performance and the same observation holds for 500 and 700. Based on this observation, we conjecture that increasing $\beta$ results in the selection of more dense pseudo labels but also introduces more hard negatives into the training. As analysed in Dense Teacher \cite{zhou2022dense}, it is helpful to include some valuable hard negatives during training, but when the proportion of hard negatives increases past a certain level, they begin to hurt the performance of the model. Generally, our method is not sensitive to hyper-parameter $\beta$.

\noindent{\bf Time cost analysis.} We report the time cost analysis of our method. Moreover, Dense Teacher \cite{zhou2022dense} and SOOD \cite{hua2023sood} are also evaluated for comparison. The results are shown in Tab 5. Our proposed method slightly increases the computational cost. In addition, our proposed method adopts teacher model for inference and thus no extra computational expense is introduced compared to the base model.

\subsection{Limitations and Discussion}

Although our method achieves satisfactory results in semi-supervised oriented object detection, its usage of the distinctive characteristics of aerial objects is still limited. Specifically, we only consider the dense distribution in our method. However, many other characteristics, such as the large scale ratio and complex background could also be considered.

\section{Conclusion}

In this paper, we find that ignoring the density of potential objects in the existing dense pseudo-label selection methods impedes performance in semi-supervised oriented object detection. To alleviate this problem, we design a simple but effective method called Density-Guided Dense Pseudo Label Selection (DDPLS) to estimate the density of potential objects and use it to guide the selection of dense pseudo label. To validate the effectiveness of our method, we have conducted extensive experiments on the DOTA-v1.5 benchmark. Compared with state-of-the-art methods, DDPLS achieves improvements on both partially and fully labeled data.

\bibliographystyle{IEEEbib}
\bibliography{strings,refs}

\end{document}